\documentclass{article} 
\usepackage{iclr2025_conference,times}

\usepackage{amsmath,amsfonts,bm}

\def\eqref#1{equation~\ref{#1}}

\def\1{\bm{1}}

\DeclareMathAlphabet{\mathsfit}{\encodingdefault}{\sfdefault}{m}{sl}
\SetMathAlphabet{\mathsfit}{bold}{\encodingdefault}{\sfdefault}{bx}{n}

\iclrfinalcopy
\usepackage{hyperref}
\usepackage{url}

\usepackage{threeparttable}
\usepackage{multirow}
\usepackage{graphicx}
\usepackage{booktabs}
\usepackage{dsfont}
\usepackage[table]{xcolor}
\usepackage{enumitem}

\renewcommand{\arraystretch}{0.8}

\usepackage{pifont}
\newcommand{\xmark}{\ding{55}}%

\title{Evaluating Inter-Column Logical Relationships in Synthetic Tabular Data Generation}

\author{
Yunbo Long\(^{1}\), Liming Xu\(^{1}\), and Alexandra Brintrup\(^{1,2}\)\\
\(^1\) Department of Engineering, University of Cambridge \\
\(^2\) The Alan Turing Institute, London \\
\texttt{\{yl892, lx249, ab702\}@cam.ac.uk}
}

\begin{document}

\maketitle

\begin{abstract}
Current evaluations of synthetic tabular data mainly focus on how well joint distributions are modeled, often overlooking the assessment of their effectiveness in preserving realistic event sequences and coherent entity relationships across columns.
This paper proposes three evaluation metrics designed to assess the preservation of logical relationships among columns in synthetic tabular data. 
We validate these metrics by assessing the performance of both classical and state-of-the-art generation methods on a real-world industrial dataset.
Experimental results reveal that existing methods often fail to rigorously maintain logical consistency (e.g., hierarchical relationships in geography or organization) and dependencies (e.g., temporal sequences or mathematical relationships), which are crucial for preserving the fine-grained realism of real-world tabular data.
Building on these insights, this study also discusses possible pathways to better capture logical relationships while modeling the distribution of synthetic tabular data. The code is available at
\href{https://github.com/Yunbo-max/TabLogicEval}
{\textcolor{red}{https://github.com/Yunbo-max/TabLogicEval}}.
\end{abstract}

\section{Introduction}
Tabular data is challenging to synthesize due to its heterogeneity, where columns can contain different variable types, exhibiting diverse distributions and complex interdependencies \citep{wang2024challenges}.
These characteristics make it difficult to accurately model the joint distribution \( P(X, Y) \) \citep{margeloiu2024tabebm} of tabular data, where \( X \) denotes the feature space and \( Y \) denotes the target variable(s).
To address these challenges, \citet{xu2019modeling} proposed GTGANs, a variant of generative adversarial netowrk (GAN) that learns the joint distribution \( P(X, Y) \) through a {\it minimax game} between a generator and a discriminator. 
While the generator produces synthetic data \( \mathbf{x}_\text{syn} = [x_1, x_2, \dots, x_n] \) by conditioning on a {\it latent} vector \citep{zhao2021ctab}, it often fails to explicitly capture specific conditional dependencies, such as \( P(x_1 \mid x_2) \), because it does not directly model the logical relationships between features.
Building on the success of diffusion models in image generation, recent work has adapted them for tabular data generation.
For example, TabDDPM \citep{kotelnikov2023TabDDPM} treats continuous and categorical features in tables separately, while TabSyn \citep{zhang2023mixed} embeds them as tokens together for diffusion processes. 
During training, those methods approximate the data distribution \( P(X_1, X_2, \dots, X_n, Y) \) by modeling the transition between noisy data at each step.  
In the forward process, noise is added {\it isotropically} to each feature according to \( q(\mathbf{x}_t | \mathbf{x}_{t-1}) = \mathcal{N}(\mathbf{x}_t; \sqrt{\alpha_t} \mathbf{x}_{t-1}, \beta_t \mathbf{I}) \), where the identity matrix (\( \mathbf{I} \)) ensures isotropic noise addition \citep{lee2023codi}. 
However, this isotropic noise assumption limits the model's ability to capture semantic dependencies among features.
Unlike images, where adjacent pixels exhibit strong spatial correlations that aid denoising \citep{nichol2021improved}, tabular data consists of heterogeneous and non-linear feature relationships without a natural ordering \citep{ruan2024language}, making it challenging for diffusion models to preserve logic dependencies.
Recently, autoregressive models, such as large language models (LLMs), have also been leveraged to approximate the joint distribution between features and target values through sequential modeling\citep{fang2024large}.  
These methods, such as GReaT \citep{borisov2022language}, transform tabular data into sequences of tokens, enabling the autoregressive models to predict the conditional probability of the  
$j$-th token given the preceding $j-1$ tokens in the permuted sequence $P(t_{i,j} \mid \Pi(t_i, k)_{1:j-1})$, where each sequence of tokens is represented as \( t_i = [t_{i,1}, t_{i,2}, \dots, t_{i,m}] \), and the sequence is randomly reordered based on a permutation vector \( k = [k_1, k_2, \dots, k_m] \).  
While LLMs capture column dependencies based on pretrained knowledge \citep{sui2024table}, they do not explicitly model the marginal distribution \( P(t_{i,j}) \), leading to {\it biased sampling} despite the introduction of feature permutations or data variability.

To evaluate the {\it fidelity} of synthetic tabular data, numerous metrics have been proposed to assess accuracy and diversity, including both low-order statistics (e.g., Density Estimation and Correlation Score \citep{zhang2023mixed}, Average Coverage Scores \citep{zein2022tabular}) and high-order statistics (e.g., $\alpha$-Precision and $\beta$-Recall \citep{alaa2022faithful}). 
However, these metrics operate at a high level and fail to evaluate whether synthetic data preserves logical relationships, such as hierarchical or semantic dependencies between features. 
This highlights the need for a more fine-grained, context-aware evaluation of multivariate dependencies.
To address this, we propose three evaluation metrics: 
Hierarchical Consistency Score (HCS), 
Multivariate Dependency Index (MDI), and 
Distributional Similarity Index (DSI).
To assess the effectiveness of these metrics in quantifying inter-column relationships, we select five representative tabular data generation methods from different categories for evaluation. 
Their performance is measured using both existing and our proposed metrics on a real-world dataset rich in logical consistency and dependency constraints. 
Experimental results validate the effectiveness of our proposed metrics and reveal the limitations of existing approaches in preserving logical relationships in synthetic tabular data.
Additionally, we discuss potential pathways to better capture logical constraints within joint distributions, paying the way for future advancements in synthetic tabular data generation.

\section{Inter-Column Relationships Evaluation Metrics}
Logical relationships inherently capture both hierarchical consistency (e.g., city \(\rightarrow\) country) and multivariate dependencies (e.g., temporal or mathematical relationships), reflecting structured interdependencies between columns.
To evaluate hierarchical consistency, we define the \textbf{Hierarchical Consistency Score (HCS)}, which quantifies how well synthetic data preserves hierarchical relationships across columns. This metric is defined as:
\begin{equation}
    \text{HCS} = \frac{1}{M \times N} \sum_{k=1}^{N} \sum_{j=1}^{M} \mathds{1}\left( (x_{i,j})_{i \in G_k} \in C_{k,j} \right),
\end{equation}
where \( M \) is the number of rows and \( N \) is the number of consistency groups, and \( x_{i,j} \) denotes the \( i \)-th attribute in the \( j \)-th row of the dataset. The set \( G_k \) refers to the \( k \)-th group of attributes (e.g., \( G_1 = \{1, 2, 3\} \)), while \( C_{k,j} \) is a set of tuples, where each tuple represents a valid combination of attribute values for the group \( G_k \) in the \( j \)-th row. The indicator function \( \mathds{1}(\cdot) \) returns 1 if the tuple \( (x_{i,j})_{i \in G_k} \) belongs to \( C_{k,j} \), and 0 otherwise.
To evaluate multivariate dependencies, we introduce the {\bf Multivariate Dependency Index} (MDI), as follows:
\begin{equation}
    \text{MDI} = \frac{1}{M \times N} \sum_{g=1}^{N} \sum_{j=1}^{M} \mathds{1} \left( D_{g,j} \right),
    \quad  D_{g,j} = \mathcal{F}(\{ x_{i,j} \mid i \in G_g \}, x_{1,j},x_{2,j} \dots, x_{n,j}).
\end{equation}
Where \( N \) is the number of dependency groups, \( M \) is the number of rows, and \( D_{g,j} \) is the Boolean dependency condition for the \( g \)-th group \( G_g \) in the \( j \)-th row. For \( i \in G_g \), \( x_{i,j} \) must satisfy the dependency function \( \mathcal{F} \) with respect to other \(n\) attributes in \( G_g \) in \( j \)-th row.
To capture multivariate dependencies in non-linear relationships where explicit dependency rules are difficult to define, we propose the {\bf Distributional Similarity Index} (DSI), which compares the log-likelihoods of Gaussian Mixture Models fitted to the each row in the synthetic dataset and the real dataset, as follows:
\begin{equation}
    \text{DSI} = \frac{1}{K} \sum_{i=1}^{K} \left( 1 - \frac{\left| \log \mathcal{L}(\hat{\mathbf{X}}^*_{\text{syn},i}) - \log \mathcal{L}(\hat{\mathbf{X}}^*_{\text{real}}) \right|}{\left| \log \mathcal{L}(\hat{\mathbf{X}}^*_{\text{real}}) \right|} \right),
\end{equation}
where \( K \) is the number of rows, \( \log \mathcal{L}(\hat{\mathbf{X}}^*_{\text{syn},i}) \) and \( \log \mathcal{L}(\hat{\mathbf{X}}^*_{\text{real}}) \) are the log-likelihood of the GMM for the \( i \)-th synthetic and real datasets separately. The GMM log-likelihood for a dataset \( \mathbf{X} \) is $\log \mathcal{L}(\mathbf{X}) = \sum_{j=1}^{n} \log \left( \sum_{k=1}^{K} \pi_k \mathcal{N}(\mathbf{x}_j \mid \boldsymbol{\mu}_k, \boldsymbol{\Sigma}_k) \right)$, where \( \pi_k \) is the mixing coefficient, and \( \mathcal{N}(\mathbf{x}_j \mid \boldsymbol{\mu}_k, \boldsymbol{\Sigma}_k) \) is the probability density of the \( j \)-th data point under the \( k \)-th Gaussian component, with mean \( \boldsymbol{\mu}_k \) and covariance \( \boldsymbol{\Sigma}_k \). DSI reflects fine-grained differences between synthetic and real data.

\section{Experiments and Discussion}

\begin{table}[t]
    \centering
    \caption{
        Results on inter-column logical relationship preservation.
        ({\bf Note}: Higher values indicate better performance. 
        The best and second-best results are highlighted in {\color{blue}\bf bold} and {\bf bold}, respectively.) 
    }
    \vspace{1em} 
    \label{tab:semantic}
    \small{
    \resizebox{\textwidth}{!}{
    \begin{tabular}{ccccccc}
      \toprule
      \multirow{2}{*}{\bf Metrics} & \multicolumn{1}{c}{\bf Interpolation-based} & \multicolumn{3}{c}{\bf Latent Space Representation-based} & \multicolumn{1}{c}{\bf LLM-based} \\  \cmidrule(l{2pt}r{2pt}){2-2} \cmidrule(l{2pt}r{2pt}){3-5}
      \cmidrule(l{2pt}r{2pt}){6-6}
         & SMOTE & CTGAN & TabDDPM & TabSyn & GReaT \\ \midrule
    Density Estimation        & {\color{blue}\bf 98.02±0.02} & 90.38±0.03 & 33.11±0.02 & \bf{96.38±0.04}                   & 89.58±0.02 \\
Correlation Score          & {\color{blue}\bf 96.21±0.62} & 74.41±0.14 & 36.78±0.01 & {\bf 94.81±0.04} & 71.00±1.16 \\
Average Coverage Scores     &  {\bf 99.41±0.08}   & 82.27±0.14    & 76.23±0.23 & {\color{blue}\bf 99.52±0.11} & 92.34±0.15 \\
$\alpha$-Precision Scores   & \bf{93.54±0.01}          & 88.90±0.14 & 0.00±0.00  & \color{blue}{\bf 98.48±0.10} & 82.18±0.15 \\
$\beta$-Recall Scores      & {\color{blue}\bf 72.21±0.13} & 1.71±0.01  & 0.00±0.00  & 22.62±0.12                   & \bf{24.05±0.14}\\ \midrule
      \rowcolor{gray!30}
      HCS & {\color{blue}\bf 98.09$\pm$0.01} & 39.23$\pm$0.03 & 16.07$\pm$0.01 & 71.63$\pm$0.08 & \textbf{98.01$\pm$0.01 } \\ 
      \rowcolor{gray!30}
      MDI & {\bf 87.03$\pm$0.03} & 38.87$\pm$0.05 & 59.08$\pm$0.05 & 68.34$\pm$0.08 & {\color{blue}\bf 97.37$\pm$0.02} \\
      \rowcolor{gray!30}
      DSI & 77.46$\pm$0.01 & 11.10$\pm$0.00 &  68.38$\pm$0.01 & {\bf83.62$\pm$0.02} & {\color{blue}\bf 85.55$\pm$0.01} \\
     \bottomrule
    \end{tabular}%
    }
}
\end{table}

We selected five representative synthetic tabular data generation methods for experiments. 
Originally introduced in early 2000s to address dataset imbalance \citep{chawla2002smote}, the interpolation-based method---SMOTE---continues to outperform many generative models \citep{margeloiu2024tabebm}.
As such, we include it as a strong baseline, alongside  the four state-of-the-art methods from different categories: CTGAN \citep{xu2019modeling}, TabDDPM \citep{kotelnikov2023TabDDPM}, TabSyn \citep{zhang2023mixed}, and GReaT \citep{borisov2022language}. 
Each method generates around 140,000 samples from the DataCo dataset\footnote{\url{https://data.mendeley.com/datasets/8gx2fvg2k6/3}} (see more about this dataset in \autoref{app:datasets}).
To ensure robustness, each evaluation was repeated ten times (see more experimental settings in \autoref{app:experiments}). 
We reported the mean and standard deviation of these ten runs in \autoref{tab:semantic}.
As shown in \autoref{tab:semantic}, SMOTE and TabSyn achieve the best overall performance in terms of low-order statistical accuracies. 
With the highest accuracy in $\alpha$-precision, TabSyn excels in accurately modeling the joint distribution among columns. 
Meanwhile, SMOTE significantly outperforms others in balancing data categories, achieving the highest $\beta$-recall scores.
However, generative model-based methods, such as CTGAN, TabDDPM, and GReaT, struggle to accurately capture the distribution when synthesizing this complex and large-scale dataset, resulting in low values for low-order statistical metrics.
Regarding inter-column relationship preservation, the results differ marginally.
For data consistency, SMOTE and GReaT achieve the highest HCS of 98.09\% and 98.01\%, respectively, outperforming all other approaches. 
For data dependency, the MDI results indicate that GReaT effectively captures temporal and mathematical dependencies, achieving approximately 97.37\% accuracy, considerably higher than the second-best method---SMOTE. 
Additionally, GReaT outperforms others in terms of DSI, however, it may generate unseen values for certain attributes, resulting in uncontrollable generation.
In comparison, latent space-based generative models (CTGAN, TabDDPM, and TabSyn) are more reliable but struggle to effectively capture inter-column logical relationships in real-world tabular data. 
See the details and examples of generated data by each method in \autoref{app:logic}.

\section{Conclusion and Research Directions}
We introduce three metrics---HCS, MDI, and DSI---for evaluating inter-column logical relationship in synthetic tabular data generation. 
Our experiments show that existing methods often fail to strictly maintain hierarchical consistency and multivariate dependencies---essential characteristics of real-world datasets. 
Our future work will focus on enhancing the preservation of inter-column logical relationships in synthetic tabular data generation.
For LLM-based methods, the column serialization format and order are crucial for the model's ability to learn the joint distribution of logically related features. 
Knowledge graphs \citep{dong2024large} or Bayesian networks \citep{ling2024mallm} would be employed to reorder tokenization sequences or restructure the serialization of columns in natural language, leveraging prior knowledge to guide the synthetic tabular data generation. 
For latent space-based methods, LLM reasoning \citep{hegselmann2023tabllm, dong2024large} can be utilized to analyze column names and descriptions, identifying semantic or logical relationships without prior knowledge. 
Additionally, inspired by CTSyn \citep{lin2024ctsyn}, grouping data by logical relationships and embedding them into a shared latent space could potentially capture inherent structures, improving joint distribution modeling.
Lastly, incorporating interpolation techniques like SMOTE may help can help balance data classes \citep{yang2024balanced}, particularly in learning minority logical relationships.
These directions are {\it worthy} to explore for designing generative models that effectively capture inter-column logical relationships in synthetic tabular data generation.



\bibliography{iclr2025_conference}
\bibliographystyle{iclr2025_conference}

\newpage
\appendix
\section{Appendix}

\subsection{Dataset}
\label{app:datasets}
DataCo is a large real-world dataset that includes complex, high-dimensional features and a wide variety of values for each feature. 
A summary of this complex dataset's characteristics is presented in \autoref{tab:exp-dataset}. 
The dataset contains around 140,000 samples designed for model training, and each method generate same number of samples for synthetic tabular data evaluation. 
Notably, certain categorical columns contain over 3,000 unique values, reflecting the dataset's complexity.

This dataset represents a series of  events related to purchasing, production, sales, and commercial distribution for an e-commerce company operating in global markets. This dataset is crucial and representative for synthetic tabular data generation in because its high-dimensional features, diverse inter-column relationships, and extensive unique values in categorical columns provide a realistic and challenging benchmark for evaluating the ability of synthetic data models to capture complex real-world patterns.

\begin{table}[h]
    \centering
    \caption{
        Statistics of the Dataco dataset. 
        ``Numerical'' represents the number of numerical columns, and ``Categorical'' stands for the number of categorical columns.
    }
    \vspace{1em} 
    \label{tab:exp-dataset}
    \small
    \begin{threeparttable}
        \scalebox{1.0}{
            \begin{tabular}{lcccccc}
                \toprule
                \textbf{Dataset} & \textbf{Rows} & \textbf{Numerical} & \textbf{Categorical} & \textbf{Train} & \textbf{Validation} & \textbf{Test} \\
                \midrule
                Dataco & 172,766 & 26 & 15 & 138,213 & 17,277 & 17,277 \\
                \bottomrule
            \end{tabular}
        }
    \end{threeparttable}
\end{table}

\subsubsection{Hierarchical Consistency}
For hierarchical consistency in the Dataco dataset, there are three sets of tuples, \( C_1 \), \( C_2 \), and \( C_3 \), which represent the sets of valid tuples for specific attribute groups in the dataset. 
Here, \( i \) represents the attribute index (e.g., the \( i \)-th attribute in the dataset), and \( j \) represents the row index (e.g., the \( j\)-th row in the dataset). 
The tuple \( \left( (x_{i,j})_{i \in G_1} \right) \) representing the \textbf{geographical information of orders} in each row \( j \) includes the following features :
\begin{itemize}[nosep]
    \item \( x_{1,j} \) (order city),
    \item \( x_{2,j} \) (order state),
    \item \( x_{3,j} \) (order country),
    \item \( x_{4,j} \) (order region),
    \item \( x_{5,j} \) (order market).
\end{itemize}
 The tuple \( \left( (x_{i,j})_{i \in G_2} \right) \) representing the \textbf{product information} in each row \( j \) includes the following features :
\begin{itemize}[nosep]
    \item \( x_{6,j} \) (category ID),
    \item \( x_{7,j} \) (category name),
    \item \( x_{8,j} \) (department ID),
    \item \( x_{9,j} \) (department name),
    \item \( x_{10,j} \) (product card ID),
    \item \( x_{11,j} \) (product category ID),
    \item \( x_{12,j} \) (product name).
\end{itemize}
 The tuple \( \left( (x_{i,j})_{i \in G_3} \right) \) representing the \textbf{geographical information of customers} in each row \( j \) includes the following features :
\begin{itemize}[nosep]
    \item \( x_{13,j} \) (customer city),
    \item \( x_{14,j} \) (customer state),
    \item \( x_{15,j} \) (customer country).
\end{itemize}

To ensure hierarchical consistency, we verify whether each tuple \( \left( (x_{i,j})_{i \in G_k} \right) \) belongs to its corresponding valid set of tuples \( C_{k,j} \), where \( k \in \mathbb{N}^+\) and \( j \) represents the row index (i.e., the \( j \)-th row in the dataset). For example, the tuple \( \left( (x_{i,1})_{i \in G_3} \right) \), where \( G_3 = \{13, 14, 15\} \) represents the combination of geographical information of customers, must belong to \( C_{3,1} \) in the first row. This validation ensures that all attribute values are consistent with their hierarchical relationships.

\subsubsection{Temporal Dependency}
Temporal dependencies are evident in the order information, where order dates and delivery times are sequentially linked. Let \( G_1 \) represent the \textbf{temporal group}, which includes \( x_1 \) as the order date and \( x_2 \) as the delivery time. The Boolean dependency condition \( D_{1,j} \) for the \( j \)-th row in \( G_1 \) is defined as: 
\[
D_{1,j}: x_{1,j} < x_{2,j},
\]
where \( x_{1,j} \) (order date) must be earlier than \( x_{2,j} \) (delivery time) to satisfy the temporal dependency.

\subsubsection{Mathematical Dependency}
Mathematical dependencies are presented in financial data. 
Let \( G_2 \) represent the \textbf{financial group}, which includes \( x_1 \) as the products quantity, \( x_2 \) as the products price, \( x_3 \) as the discount rate, \( x_4 \) as the discount value, \( x_5 \) as the original price, and \( x_6 \) as the sales price. The Boolean dependency conditions \( D_{1,j} \), \( D_{2,j} \), and \( D_{3,j} \) for the \( j \)-th row in \( G_2 \) are defined as:
\[
D_{1,j}: x_{5,j} = x_{1,j} \times x_{2,j},
\]
\[
D_{2,j}: x_{5j} = x_{1,j} \times x_{2,j} \times x_{3,j} ,
\]
\[
D_{3,j}: x_{4,j} = x_{7,j} - x_{5,j},
\]
where the discount value should be equal to the product of the product price and the discount rate. The sales price should be calculated as the original price minus the discount value. Additionally, the original price should be determined by multiplying the product's quantity by its unit price.

\subsection{Experiments Settings}
\label{app:experiments}
To ensure reproducibility and fairness, all experiments are conducted on an open-source industrial dataset---DataCo. 
For SMOTE, synthetic data is sampled ten times to ensure robustness and account for variability in the interpolation-based generation process. 
For other methods (CTGAN, TabDDPM, TabSyn, and GReaT), models are trained on 80\% of the dataset, with 10\% used for validation and 10\% for testing, following the default hyperparameter settings from their respective papers. 
The trained models are then used to generate synthetic tabular data. 
Each method is evaluated ten times, and the results are reported as the mean and standard deviation to account for variability.

To comprehensively evaluate the quality of synthetic data, each method is assessed using a set of metrics designed to measure different aspects of data fidelity and utility. 
These metrics include:
\begin{itemize}[nosep]
    \item {\bf Hierarchical Consistency Score (HCS)}: 
    Measures the preservation of hierarchical relationships in the data.
    
    \item {\bf Multivariate Dependency Index (MDI)}: 
    Quantifies the preservation of multivariate dependencies between features.
    
    \item {\bf Distributional Similarity Index (DSI)}:
    Evaluates the overall similarity between the synthetic and real data distributions, where we embed both categorical and numerical features into a shared continuous space. 
\end{itemize}

Additionally, we also adopted existing statistical metrics for evaluation:
\begin{itemize}[nosep]
    \item \textbf{Density Estimation}: Measures how well the synthetic data matches the probability density of the real data, with higher values indicating better alignment \citep{zhang2023mixed}.
    
    \item \textbf{Correlation Score}: Quantifies the preservation of pairwise correlations between features, with higher scores reflecting better retention of feature relationships \citep{zhang2023mixed}.
    
    \item \textbf{Average Coverage Score}: Evaluates the extent to which the synthetic data covers the range of values in the real data, with higher scores indicating better diversity \citep{zein2022tabular}.
    
    \item \textbf{\(\alpha\)-Precision Score}: Evaluates the fidelity of synthetic data by determining whether each synthetic example originates from the real-data distribution, providing a measure of how well the synthetic data aligns with the true underlying data structure.\citep{alaa2022faithful}.
    
    \item \textbf{\(\beta\)-Recall Score}: Evaluates how well the synthetic data captures the real data distribution, with higher values indicating better representation \citep{alaa2022faithful}.
\end{itemize}
 
As current experiments were conducted over a single dataset, our future work will validate these three metrics (HCS, MDI, and DSI) on broader datasets to ensure their generalizability and robustness across different domains.

\subsection{Additional Results on Logical Relationships Preservation}
\label{app:logic}
To further examine the generated tabular data, we randomly select 10 rows from the generated dataset produced by each method to illustrate how effectively they preserve logical relationships. 
Specifically, we examine three types of relationships: mathematical dependencies, geographical hierarchies, and temporal sequences.

\paragraph{Mathematical Dependencies}
For mathematical dependency, as shown in \autoref{tab:maths}, it is evident that CTGAN and TabDDPM produce numerous errors in maintaining these relationships, with none of randomly selected samples preserve mathematic dependency. 
In contrast, TabSyn successfully captures the mathematical relationships for all sampled rows. 
GReaT performs fairly good, preserving most of the mathematical relationships completely across the ten samples, followed closely by SMOTE.

\paragraph{Geographical Hierarchies}
For geographical relationship preservation, as shown in \autoref{tab:hierachical}, CTGAN and TabDDPM fail to maintain any correct consistency in the logical chains. 
TabSyn, however, preserves the hierarchical consistency for some samples. 
GReaT and SMOTE achieve nearly 100\% accuracy in maintaining geographical consistency, though GReaT exhibits a small number of errors, including generating incorrect names for order cities.

\paragraph{Temporal Relationships}
For temporal relationship(see \autoref{tab:temperal}), latent space-based methods (e.g., CTGAN, TabDDPM) preserve only about half of the correct temporal relationships, with results appearing somewhat random. 
In contrast, GReaT and SMOTE achieve near-perfect performance in preserving temporal relationships, although minor errors occur when the delivery and order dates are very close.

These results demonstrate that GReaT and SMOTE outperform other methods in preserving logical relationships, with GReaT showing slight inconsistencies in geographical and mathematical relationships. 
TabSyn performs moderately well in capturing temporal dependencies but struggles with geographical and mathematical consistency. 
In contrast, CTGAN and TabDDPM exhibit significant limitations across all types of logical relationships.

\renewcommand{\arraystretch}{1.2} 
\begin{table}[t]
\centering
\caption{Mathmatical dependency preservation in synthetic tabular data}
\vspace{1em} 
\label{tab:maths}
\resizebox{\textwidth}{!}{%
\begin{tabular}{cccccccc|}
\hline
\multicolumn{1}{|c}{\multirow{3}{*}{\textbf{\begin{tabular}[c]{@{}c@{}}Tabular Data \\ Generation \\ Methods\end{tabular}}}} &
  \multicolumn{6}{c}{\textbf{Group 1(Financial Data)}} &
  \multirow{3}{*}{\textbf{\begin{tabular}[c]{@{}c@{}}Preserved\end{tabular}}} \\ \cline{2-7}
\multicolumn{1}{|c}{} &
  \multicolumn{1}{c}{\textbf{Attribute 1}} &
  \multicolumn{1}{c}{\textbf{Attribute 2}} &
  \multicolumn{1}{c}{\textbf{Attribute 3}} &
  \multicolumn{1}{c}{\textbf{Attribute 4}} &
  \multicolumn{1}{c}{\textbf{Attribute 5}} &
  \multicolumn{1}{c}{\textbf{Attribute 6}} &
   \\ \cline{2-7}
\multicolumn{1}{|c}{} &
  \multicolumn{1}{c}{\textbf{Quantity}} &
  \multicolumn{1}{c}{\textbf{Product Price}} &
  \multicolumn{1}{c}{\textbf{Discount Rate}} &
  \multicolumn{1}{c}{\textbf{Discount Values}} &
  \multicolumn{1}{c}{\textbf{Original Price}} &
  \multicolumn{1}{c}{\textbf{Sales Price}} &
   \\ \hline\hline
\multicolumn{1}{|c}{\multirow{3}{*}{\textbf{Original Tables}}} &
  \multicolumn{1}{c}{5} &
  \multicolumn{1}{c}{49.98} &
  \multicolumn{1}{c}{0.09} &
  \multicolumn{1}{c}{22.49} &
  \multicolumn{1}{c}{249.90} &
  \multicolumn{1}{c}{227.41} &
  \cellcolor{yellow}\textbf{\checkmark} \\ \cline{2-8} 
\multicolumn{1}{|c}{} &
  \multicolumn{1}{c}{2} &
  \multicolumn{1}{c}{49.98} &
  \multicolumn{1}{c}{0.04} &
  \multicolumn{1}{c}{4.00} &
  \multicolumn{1}{c}{99.96} &
  \multicolumn{1}{c}{95.96} &
  \cellcolor{yellow}\textbf{\checkmark} \\ \cline{2-8} 
\multicolumn{1}{|c}{} &
  \multicolumn{1}{c}{1} &
  \multicolumn{1}{c}{129.99} &
  \multicolumn{1}{c}{0.05} &
  \multicolumn{1}{c}{6.50} &
  \multicolumn{1}{c}{129.99} &
  \multicolumn{1}{c}{123.49} &
  \cellcolor{yellow}\textbf{\checkmark} \\ \hline\hline
\multicolumn{1}{|c}{\multirow{10}{*}{\textbf{CTGAN}}} &
  \multicolumn{1}{c}{5} &
  \multicolumn{1}{c}{48.40} &
  \multicolumn{1}{c}{0.02} &
  \multicolumn{1}{c}{2.53} &
  \multicolumn{1}{c}{191.56} &
  \multicolumn{1}{c}{463.84} &
  \textbf{\xmark} \\ \cline{2-8} 
\multicolumn{1}{|c}{} &
  \multicolumn{1}{c}{1} &
  \multicolumn{1}{c}{199.83} &
  \multicolumn{1}{c}{0.04} &
  \multicolumn{1}{c}{8.63} &
  \multicolumn{1}{c}{201.26} &
  \multicolumn{1}{c}{204.72} &
  \textbf{\xmark} \\ \cline{2-8} 
\multicolumn{1}{|c}{} &
  \multicolumn{1}{c}{4} &
  \multicolumn{1}{c}{297.48} &
  \multicolumn{1}{c}{0.15} &
  \multicolumn{1}{c}{46.76} &
  \multicolumn{1}{c}{200.08} &
  \multicolumn{1}{c}{198.92} &
  \textbf{\xmark} \\ \cline{2-8} 
\multicolumn{1}{|c}{} &
  \multicolumn{1}{c}{4} &
  \multicolumn{1}{c}{57.67} &
  \multicolumn{1}{c}{0.02} &
  \multicolumn{1}{c}{0.20} &
  \multicolumn{1}{c}{249.58} &
  \multicolumn{1}{c}{192.38} &
  \textbf{\xmark} \\ \cline{2-8} 
\multicolumn{1}{|c}{} &
  \multicolumn{1}{c}{5} &
  \multicolumn{1}{c}{46.77} &
  \multicolumn{1}{c}{0.15} &
  \multicolumn{1}{c}{20.41} &
  \multicolumn{1}{c}{254.27} &
  \multicolumn{1}{c}{239.82} &
  \textbf{\xmark} \\ \cline{2-8} 
\multicolumn{1}{|c}{} &
  \multicolumn{1}{c}{1} &
  \multicolumn{1}{c}{302.36} &
  \multicolumn{1}{c}{0.17} &
  \multicolumn{1}{c}{88.55} &
  \multicolumn{1}{c}{301.53} &
  \multicolumn{1}{c}{292.56} &
  \textbf{\xmark} \\ \cline{2-8} 
\multicolumn{1}{|c}{} &
  \multicolumn{1}{c}{1} &
  \multicolumn{1}{c}{129.02} &
  \multicolumn{1}{c}{0.04} &
  \multicolumn{1}{c}{0.57} &
  \multicolumn{1}{c}{131.43} &
  \multicolumn{1}{c}{100.26} &
  \textbf{\xmark} \\ \cline{2-8} 
\multicolumn{1}{|c}{} &
  \multicolumn{1}{c}{1} &
  \multicolumn{1}{c}{402.32} &
  \multicolumn{1}{c}{0.03} &
  \multicolumn{1}{c}{7.43} &
  \multicolumn{1}{c}{200.15} &
  \multicolumn{1}{c}{378.42} &
  \textbf{\xmark} \\ \cline{2-8} 
\multicolumn{1}{|c}{} &
  \multicolumn{1}{c}{2} &
  \multicolumn{1}{c}{52.56} &
  \multicolumn{1}{c}{0.16} &
  \multicolumn{1}{c}{3.34} &
  \multicolumn{1}{c}{100.02} &
  \multicolumn{1}{c}{100.01} &
  \textbf{\xmark} \\ \cline{2-8} 
\multicolumn{1}{|c}{} &
  \multicolumn{1}{c}{2} &
  \multicolumn{1}{c}{53.91} &
  \multicolumn{1}{c}{0.10} &
  \multicolumn{1}{c}{8.71} &
  \multicolumn{1}{c}{23.39} &
  \multicolumn{1}{c}{151.86} &
  \textbf{\xmark} \\ \hline\hline
\multicolumn{1}{|c}{\multirow{10}{*}{\textbf{TabDDPM}}} &
  \multicolumn{1}{c}{1} &
  \multicolumn{1}{c}{1999.99} &
  \multicolumn{1}{c}{0} &
  \multicolumn{1}{c}{500} &
  \multicolumn{1}{c}{1999.99} &
  \multicolumn{1}{c}{1939.99} &
  \textbf{\xmark} \\ \cline{2-8} 
\multicolumn{1}{|c}{} &
  \multicolumn{1}{c}{5} &
  \multicolumn{1}{c}{9.99} &
  \multicolumn{1}{c}{0} &
  \multicolumn{1}{c}{500} &
  \multicolumn{1}{c}{9.99} &
  \multicolumn{1}{c}{7.99} &
  \textbf{\xmark} \\ \cline{2-8} 
\multicolumn{1}{|c}{} &
  \multicolumn{1}{c}{5} &
  \multicolumn{1}{c}{1999.99} &
  \multicolumn{1}{c}{0.25} &
  \multicolumn{1}{c}{500} &
  \multicolumn{1}{c}{1999.99} &
  \multicolumn{1}{c}{1939.99} &
  \textbf{\xmark} \\ \cline{2-8} 
\multicolumn{1}{|c}{} &
  \multicolumn{1}{c}{5} &
  \multicolumn{1}{c}{1999.99} &
  \multicolumn{1}{c}{0} &
  \multicolumn{1}{c}{500} &
  \multicolumn{1}{c}{9.99} &
  \multicolumn{1}{c}{1939.99} &
  \textbf{\xmark} \\ \cline{2-8} 
\multicolumn{1}{|c}{} &
  \multicolumn{1}{c}{1} &
  \multicolumn{1}{c}{9.99} &
  \multicolumn{1}{c}{0} &
  \multicolumn{1}{c}{500} &
  \multicolumn{1}{c}{1999.99} &
  \multicolumn{1}{c}{1939.99} &
  \textbf{\xmark} \\ \cline{2-8} 
\multicolumn{1}{|c}{} &
  \multicolumn{1}{c}{1} &
  \multicolumn{1}{c}{9.99} &
  \multicolumn{1}{c}{0} &
  \multicolumn{1}{c}{500} &
  \multicolumn{1}{c}{1999.99} &
  \multicolumn{1}{c}{1939.99} &
  \textbf{\xmark} \\ \cline{2-8} 
\multicolumn{1}{|c}{} &
  \multicolumn{1}{c}{1} &
  \multicolumn{1}{c}{9.99} &
  \multicolumn{1}{c}{0} &
  \multicolumn{1}{c}{0} &
  \multicolumn{1}{c}{9.99} &
  \multicolumn{1}{c}{1939.99} &
  \textbf{\xmark} \\ \cline{2-8} 
\multicolumn{1}{|c}{} &
  \multicolumn{1}{c}{1} &
  \multicolumn{1}{c}{1999.99} &
  \multicolumn{1}{c}{0} &
  \multicolumn{1}{c}{500} &
  \multicolumn{1}{c}{1999.99} &
  \multicolumn{1}{c}{7.99} &
  \textbf{\xmark} \\ \cline{2-8} 
\multicolumn{1}{|c}{} &
  \multicolumn{1}{c}{5} &
  \multicolumn{1}{c}{1999.99} &
  \multicolumn{1}{c}{0} &
  \multicolumn{1}{c}{500} &
  \multicolumn{1}{c}{9.99} &
  \multicolumn{1}{c}{7.99} &
  \textbf{\xmark} \\ \cline{2-8} 
\multicolumn{1}{|c}{} &
  \multicolumn{1}{c}{1} &
  \multicolumn{1}{c}{9.99} &
  \multicolumn{1}{c}{0} &
  \multicolumn{1}{c}{500} &
  \multicolumn{1}{c}{9.99} &
  \multicolumn{1}{c}{7.99} &
  \textbf{\xmark} \\ \hline\hline
\multicolumn{1}{|c}{\multirow{10}{*}{\textbf{TabSyn}}} &
  \multicolumn{1}{c}{1} &
  \multicolumn{1}{c}{299.98} &
  \multicolumn{1}{c}{0.01} &
  \multicolumn{1}{c}{3} &
  \multicolumn{1}{c}{299.98} &
  \multicolumn{1}{c}{299.96} &
  \textbf{\xmark} \\ \cline{2-8} 
\multicolumn{1}{|c}{} &
  \multicolumn{1}{c}{1} &
  \multicolumn{1}{c}{129.99} &
  \multicolumn{1}{c}{0.06} &
  \multicolumn{1}{c}{7.2} &
  \multicolumn{1}{c}{129.99} &
  \multicolumn{1}{c}{123.66} &
  \textbf{\xmark} \\ \cline{2-8} 
\multicolumn{1}{|c}{} &
  \multicolumn{1}{c}{1} &
  \multicolumn{1}{c}{199.99} &
  \multicolumn{1}{c}{0.04} &
  \multicolumn{1}{c}{8} &
  \multicolumn{1}{c}{199.99} &
  \multicolumn{1}{c}{191.97} &
  \cellcolor{yellow}\textbf{\checkmark} \\ \cline{2-8} 
\multicolumn{1}{|c}{} &
  \multicolumn{1}{c}{1} &
  \multicolumn{1}{c}{251.37} &
  \multicolumn{1}{c}{0.04} &
  \multicolumn{1}{c}{10.72} &
  \multicolumn{1}{c}{299.95} &
  \multicolumn{1}{c}{254.67} &
  \textbf{\xmark} \\ \cline{2-8} 
\multicolumn{1}{|c}{} &
  \multicolumn{1}{c}{4} &
  \multicolumn{1}{c}{99.99} &
  \multicolumn{1}{c}{0.03} &
  \multicolumn{1}{c}{10.82} &
  \multicolumn{1}{c}{399.96} &
  \multicolumn{1}{c}{387.98} &
  \cellcolor{yellow}\textbf{\checkmark} \\ \cline{2-8} 
\multicolumn{1}{|c}{} &
  \multicolumn{1}{c}{3} &
  \multicolumn{1}{c}{50} &
  \multicolumn{1}{c}{0.05} &
  \multicolumn{1}{c}{7.46} &
  \multicolumn{1}{c}{179.96} &
  \multicolumn{1}{c}{139.50} &
  \textbf{\xmark} \\ \cline{2-8} 
\multicolumn{1}{|c}{} &
  \multicolumn{1}{c}{5} &
  \multicolumn{1}{c}{99.99} &
  \multicolumn{1}{c}{0.13} &
  \multicolumn{1}{c}{70.68} &
  \multicolumn{1}{c}{499.95} &
  \multicolumn{1}{c}{427.09} &
  \textbf{\xmark} \\ \cline{2-8} 
\multicolumn{1}{|c}{} &
  \multicolumn{1}{c}{5} &
  \multicolumn{1}{c}{50} &
  \multicolumn{1}{c}{0.05} &
  \multicolumn{1}{c}{12.5} &
  \multicolumn{1}{c}{250} &
  \multicolumn{1}{c}{239.98} &
  \textbf{\xmark} \\ \cline{2-8} 
\multicolumn{1}{|c}{} &
  \multicolumn{1}{c}{2} &
  \multicolumn{1}{c}{50} &
  \multicolumn{1}{c}{0.25} &
  \multicolumn{1}{c}{24.60} &
  \multicolumn{1}{c}{100} &
  \multicolumn{1}{c}{73.19} &
  \textbf{\xmark} \\ \cline{2-8} 
\multicolumn{1}{|c}{} &
  \multicolumn{1}{c}{1} &
  \multicolumn{1}{c}{50} &
  \multicolumn{1}{c}{0.09} &
  \multicolumn{1}{c}{9} &
  \multicolumn{1}{c}{100} &
  \multicolumn{1}{c}{92.93} &
  \textbf{\xmark} \\ \hline\hline
\multicolumn{1}{|c}{\multirow{10}{*}{\textbf{GReaT}}} &
  \multicolumn{1}{c}{3} &
  \multicolumn{1}{c}{99.99} &
  \multicolumn{1}{c}{0.03} &
  \multicolumn{1}{c}{9} &
  \multicolumn{1}{c}{299.97} &
  \multicolumn{1}{c}{290.97} &
  \cellcolor{yellow}\textbf{\checkmark} \\ \cline{2-8} 
\multicolumn{1}{|c}{} &
  \multicolumn{1}{c}{4} &
  \multicolumn{1}{c}{50} &
  \multicolumn{1}{c}{0.15} &
  \multicolumn{1}{c}{30} &
  \multicolumn{1}{c}{200} &
  \multicolumn{1}{c}{170} &
  \cellcolor{yellow}\textbf{\checkmark} \\ \cline{2-8} 
\multicolumn{1}{|c}{} &
  \multicolumn{1}{c}{1} &
  \multicolumn{1}{c}{199.99} &
  \multicolumn{1}{c}{0.09} &
  \multicolumn{1}{c}{18} &
  \multicolumn{1}{c}{199.99} &
  \multicolumn{1}{c}{181.99} &
  \cellcolor{yellow}\textbf{\checkmark} \\ \cline{2-8} 
\multicolumn{1}{|c}{} &
  \multicolumn{1}{c}{1} &
  \multicolumn{1}{c}{129.99} &
  \multicolumn{1}{c}{0.13} &
  \multicolumn{1}{c}{16.90} &
  \multicolumn{1}{c}{129.99} &
  \multicolumn{1}{c}{113.09} &
  \cellcolor{yellow}\textbf{\checkmark} \\ \cline{2-8} 
\multicolumn{1}{|c}{} &
  \multicolumn{1}{c}{1} &
  \multicolumn{1}{c}{50} &
  \multicolumn{1}{c}{0.16} &
  \multicolumn{1}{c}{8} &
  \multicolumn{1}{c}{50} &
  \multicolumn{1}{c}{42} &
  \cellcolor{yellow}\textbf{\checkmark} \\ \cline{2-8} 
\multicolumn{1}{|c}{} &
  \multicolumn{1}{c}{1} &
  \multicolumn{1}{c}{199.99} &
  \multicolumn{1}{c}{0.04} &
  \multicolumn{1}{c}{8} &
  \multicolumn{1}{c}{199.99} &
  \multicolumn{1}{c}{191.99} &
  \cellcolor{yellow}\textbf{\checkmark} \\ \cline{2-8} 
\multicolumn{1}{|c}{} &
  \multicolumn{1}{c}{4} &
  \multicolumn{1}{c}{49.98} &
  \multicolumn{1}{c}{0.10} &
  \multicolumn{1}{c}{19.99} &
  \multicolumn{1}{c}{199.92} &
  \multicolumn{1}{c}{179.93} &
  \cellcolor{yellow}\textbf{\checkmark} \\ \cline{2-8} 
\multicolumn{1}{|c}{} &
  \multicolumn{1}{c}{2} &
  \multicolumn{1}{c}{49.98} &
  \multicolumn{1}{c}{0.04} &
  \multicolumn{1}{c}{12} &
  \multicolumn{1}{c}{99.96} &
  \multicolumn{1}{c}{95.96} &
  \textbf{\xmark} \\ \cline{2-8} 
\multicolumn{1}{|c}{} &
  \multicolumn{1}{c}{1} &
  \multicolumn{1}{c}{399.98} &
  \multicolumn{1}{c}{0.03} &
  \multicolumn{1}{c}{12} &
  \multicolumn{1}{c}{399.98} &
  \multicolumn{1}{c}{387.98} &
  \cellcolor{yellow}\textbf{\checkmark} \\ \cline{2-8} 
\multicolumn{1}{|c}{} &
  \multicolumn{1}{c}{2} &
  \multicolumn{1}{c}{50} &
  \multicolumn{1}{c}{0.07} &
  \multicolumn{1}{c}{7} &
  \multicolumn{1}{c}{100} &
  \multicolumn{1}{c}{93} &
  \cellcolor{yellow}\textbf{\checkmark} \\ \hline\hline
\multicolumn{1}{|c}{\multirow{10}{*}{\textbf{SMOTE}}} &
  \multicolumn{1}{c}{1} &
  \multicolumn{1}{c}{129.99} &
  \multicolumn{1}{c}{0.04} &
  \multicolumn{1}{c}{5.40} &
  \multicolumn{1}{c}{129.99} &
  \multicolumn{1}{c}{124.22} &
  \textbf{\xmark} \\ \cline{2-8} 
\multicolumn{1}{|c}{} &
  \multicolumn{1}{c}{1} &
  \multicolumn{1}{c}{50} &
  \multicolumn{1}{c}{0.16} &
  \multicolumn{1}{c}{8} &
  \multicolumn{1}{c}{50} &
  \multicolumn{1}{c}{42} &
  \cellcolor{yellow}\textbf{\checkmark} \\ \cline{2-8} 
\multicolumn{1}{|c}{} &
  \multicolumn{1}{c}{1} &
  \multicolumn{1}{c}{49.98} &
  \multicolumn{1}{c}{0.13} &
  \multicolumn{1}{c}{4.80} &
  \multicolumn{1}{c}{39.99} &
  \multicolumn{1}{c}{34.84} &
  \textbf{\xmark} \\ \cline{2-8} 
\multicolumn{1}{|c}{} &
  \multicolumn{1}{c}{5} &
  \multicolumn{1}{c}{17.99} &
  \multicolumn{1}{c}{0.02} &
  \multicolumn{1}{c}{1.60} &
  \multicolumn{1}{c}{89.95} &
  \multicolumn{1}{c}{87.97} &
  \textbf{\xmark} \\ \cline{2-8} 
\multicolumn{1}{|c}{} &
  \multicolumn{1}{c}{1} &
  \multicolumn{1}{c}{199.99} &
  \multicolumn{1}{c}{0} &
  \multicolumn{1}{c}{0} &
  \multicolumn{1}{c}{199.99} &
  \multicolumn{1}{c}{199.99} &
  \cellcolor{yellow}\cellcolor{yellow}\cellcolor{yellow}\textbf{\checkmark} \\ \cline{2-8} 
\multicolumn{1}{|c}{} &
  \multicolumn{1}{c}{1} &
  \multicolumn{1}{c}{24.99} &
  \multicolumn{1}{c}{0.09} &
  \multicolumn{1}{c}{2} &
  \multicolumn{1}{c}{24.99} &
  \multicolumn{1}{c}{21.65} &
  \textbf{\xmark} \\ \cline{2-8} 
\multicolumn{1}{|c}{} &
  \multicolumn{1}{c}{1} &
  \multicolumn{1}{c}{210.85} &
  \multicolumn{1}{c}{0} &
  \multicolumn{1}{c}{0} &
  \multicolumn{1}{c}{210.85} &
  \multicolumn{1}{c}{210.85} &
  \cellcolor{yellow}\cellcolor{yellow}\textbf{\checkmark} \\ \cline{2-8} 
\multicolumn{1}{|c}{} &
  \multicolumn{1}{c}{1} &
  \multicolumn{1}{c}{50} &
  \multicolumn{1}{c}{0.25} &
  \multicolumn{1}{c}{12.5} &
  \multicolumn{1}{c}{50} &
  \multicolumn{1}{c}{37.5} &
  \cellcolor{yellow}\cellcolor{yellow}\textbf{\checkmark} \\ \cline{2-8} 
\multicolumn{1}{|c}{} &
  \multicolumn{1}{c}{3} &
  \multicolumn{1}{c}{79.99} &
  \multicolumn{1}{c}{0.13} &
  \multicolumn{1}{c}{30} &
  \multicolumn{1}{c}{249.90} &
  \multicolumn{1}{c}{217.42} &
  \textbf{\xmark} \\ \cline{2-8} 
\multicolumn{1}{|c}{} &
  \multicolumn{1}{c}{1} &
  \multicolumn{1}{c}{129.99} &
  \multicolumn{1}{c}{0.03} &
  \multicolumn{1}{c}{3.53} &
  \multicolumn{1}{c}{129.99} &
  \multicolumn{1}{c}{126.09} &
  \textbf{\xmark} \\ \hline
\end{tabular}%
}
\end{table}

\renewcommand{\arraystretch}{1.2} 
\begin{table}[t]
\centering
\caption{Hierachical consistency (geographical relationships) preservation in synthetic tabular data}
\vspace{1em} 
\label{tab:hierachical}
\resizebox{1.0\textwidth}{!}{%
\begin{tabular}{|ccccccc|}
\hline
\multicolumn{1}{|c}{\multirow{3}{*}{\textbf{Methods}}} &
  \multicolumn{5}{c}{\textbf{Group 1(Geographical Data)}} &
  \multirow{3}{*}{\textbf{\begin{tabular}[c]{@{}c@{}}Preserved\end{tabular}}} \\ \cline{2-6}
\multicolumn{1}{|c}{} &
  \multicolumn{1}{c}{\textbf{Attribute 1}} &
  \multicolumn{1}{c}{\textbf{Attribute 2}} &
  \multicolumn{1}{c}{\textbf{Attribute 3}} &
  \multicolumn{1}{c}{\textbf{Attribute 4}} &
  \multicolumn{1}{c}{\textbf{Attribute 5}} &
   \\ \cline{2-6}
\multicolumn{1}{|c}{} &
  \multicolumn{1}{c}{\textbf{Order City}} &
  \multicolumn{1}{c}{\textbf{Order State}} &
  \multicolumn{1}{c}{\textbf{Order Country}} &
  \multicolumn{1}{c}{\textbf{Order Region}} &
  \multicolumn{1}{c}{\textbf{Order Market}} &
   \\ \hline\hline
\multicolumn{1}{|c}{\multirow{3}{*}{\textbf{Original Tables}}} &
  \multicolumn{1}{c}{Providence} &
  \multicolumn{1}{c}{Rhode Island} &
  \multicolumn{1}{c}{United States} &
  \multicolumn{1}{c}{East of USA} &
  \multicolumn{1}{c}{USCA} &
  \cellcolor{yellow}\cellcolor{yellow}\textbf{\checkmark} \\ \cline{2-7} 
\multicolumn{1}{|c}{} &
  \multicolumn{1}{c}{Porirua} &
  \multicolumn{1}{c}{Wellington} &
  \multicolumn{1}{c}{New Zealand} &
  \multicolumn{1}{c}{Oceania} &
  \multicolumn{1}{c}{Pacific Asia} &
  \cellcolor{yellow}\cellcolor{yellow}\textbf{\checkmark} \\ \cline{2-7} 
\multicolumn{1}{|c}{} &
  \multicolumn{1}{c}{Tegucigalpa} &
  \multicolumn{1}{c}{Francisco Morazan} &
  \multicolumn{1}{c}{Honduras} &
  \multicolumn{1}{c}{Central America} &
  \multicolumn{1}{c}{LATAM} &
  \cellcolor{yellow}\cellcolor{yellow}\textbf{\checkmark} \\ \hline\hline
\multicolumn{1}{|c}{\multirow{10}{*}{\textbf{CTGAN}}} &
  \multicolumn{1}{c}{Aurangabad} &
  \multicolumn{1}{c}{Yalova} &
  \multicolumn{1}{c}{Nicaragua} &
  \multicolumn{1}{c}{Central America} &
  \multicolumn{1}{c}{LATAM} &
  \textbf{\xmark} \\ \cline{2-7} 
\multicolumn{1}{|c}{} &
  \multicolumn{1}{c}{Rustenburg} &
  \multicolumn{1}{c}{Sao Paulo} &
  \multicolumn{1}{c}{Netherlands} &
  \multicolumn{1}{c}{Western Europe} &
  \multicolumn{1}{c}{Europe} &
  \textbf{\xmark} \\ \cline{2-7} 
\multicolumn{1}{|c}{} &
  \multicolumn{1}{c}{Rasht} &
  \multicolumn{1}{c}{Maluku} &
  \multicolumn{1}{c}{Mexico} &
  \multicolumn{1}{c}{Central America} &
  \multicolumn{1}{c}{LATAM} &
  \textbf{\xmark} \\ \cline{2-7} 
\multicolumn{1}{|c}{} &
  \multicolumn{1}{c}{San Pedro} &
  \multicolumn{1}{c}{Kano} &
  \multicolumn{1}{c}{Morocco} &
  \multicolumn{1}{c}{West Africa} &
  \multicolumn{1}{c}{Africa} &
  \textbf{\xmark} \\ \cline{2-7} 
\multicolumn{1}{|c}{} &
  \multicolumn{1}{c}{Medan} &
  \multicolumn{1}{c}{Victoria} &
  \multicolumn{1}{c}{Australia} &
  \multicolumn{1}{c}{Oceania} &
  \multicolumn{1}{c}{Pacific Asia} &
  \textbf{\xmark} \\ \cline{2-7} 
\multicolumn{1}{|c}{} &
  \multicolumn{1}{c}{Birobidzhan} &
  \multicolumn{1}{c}{Alsace-Champagene-Ardenne} &
  \multicolumn{1}{c}{France} &
  \multicolumn{1}{c}{Western Europe} &
  \multicolumn{1}{c}{Europe} &
  \textbf{\xmark} \\ \cline{2-7} 
\multicolumn{1}{|c}{} &
  \multicolumn{1}{c}{Lagos} &
  \multicolumn{1}{c}{Kinshasa} &
  \multicolumn{1}{c}{Turkmenistan} &
  \multicolumn{1}{c}{East Africa} &
  \multicolumn{1}{c}{Africa} &
  \textbf{\xmark} \\ \cline{2-7} 
\multicolumn{1}{|c}{} &
  \multicolumn{1}{c}{Yakarta} &
  \multicolumn{1}{c}{Punjab} &
  \multicolumn{1}{c}{India} &
  \multicolumn{1}{c}{Southeast Asia} &
  \multicolumn{1}{c}{Pacific Asia} &
  \textbf{\xmark} \\ \cline{2-7} 
\multicolumn{1}{|c}{} &
  \multicolumn{1}{c}{Roermond} &
  \multicolumn{1}{c}{Mersin} &
  \multicolumn{1}{c}{Turkey} &
  \multicolumn{1}{c}{West Asia} &
  \multicolumn{1}{c}{Pacific Asia} &
  \textbf{\xmark} \\ \cline{2-7} 
\multicolumn{1}{|c}{} &
  \multicolumn{1}{c}{Houston} &
  \multicolumn{1}{c}{Auckland} &
  \multicolumn{1}{c}{Mexico} &
  \multicolumn{1}{c}{Central America} &
  \multicolumn{1}{c}{LATAM} &
  \textbf{\xmark} \\ \hline\hline
\multicolumn{1}{|c}{\multirow{10}{*}{\textbf{TabDDPM}}} &
  \multicolumn{1}{c}{Detmold} &
  \multicolumn{1}{c}{Zagrebacka} &
  \multicolumn{1}{c}{Detmold} &
  \multicolumn{1}{c}{Southeast Asia} &
  \multicolumn{1}{c}{LATAM} &
  \textbf{\xmark} \\ \cline{2-7} 
\multicolumn{1}{|c}{} &
  \multicolumn{1}{c}{Detmold} &
  \multicolumn{1}{c}{Zagrebacka} &
  \multicolumn{1}{c}{Detmold} &
  \multicolumn{1}{c}{Southeast Asia} &
  \multicolumn{1}{c}{LATAM} &
  \textbf{\xmark} \\ \cline{2-7} 
\multicolumn{1}{|c}{} &
  \multicolumn{1}{c}{Detmold} &
  \multicolumn{1}{c}{Zagrebacka} &
  \multicolumn{1}{c}{Detmold} &
  \multicolumn{1}{c}{Southeast Asia} &
  \multicolumn{1}{c}{LATAM} &
  \textbf{\xmark} \\ \cline{2-7} 
\multicolumn{1}{|c}{} &
  \multicolumn{1}{c}{Detmold} &
  \multicolumn{1}{c}{Zagrebacka} &
  \multicolumn{1}{c}{Detmold} &
  \multicolumn{1}{c}{Southeast Asia} &
  \multicolumn{1}{c}{LATAM} &
  \textbf{\xmark} \\ \cline{2-7} 
\multicolumn{1}{|c}{} &
  \multicolumn{1}{c}{Detmold} &
  \multicolumn{1}{c}{Zagrebacka} &
  \multicolumn{1}{c}{Detmold} &
  \multicolumn{1}{c}{Southeast Asia} &
  \multicolumn{1}{c}{LATAM} &
  \textbf{\xmark} \\ \cline{2-7} 
\multicolumn{1}{|c}{} &
  \multicolumn{1}{c}{Detmold} &
  \multicolumn{1}{c}{Zagrebacka} &
  \multicolumn{1}{c}{Detmold} &
  \multicolumn{1}{c}{Southeast Asia} &
  \multicolumn{1}{c}{LATAM} &
  \textbf{\xmark} \\ \cline{2-7} 
\multicolumn{1}{|c}{} &
  \multicolumn{1}{c}{Detmold} &
  \multicolumn{1}{c}{Zagrebacka} &
  \multicolumn{1}{c}{Detmold} &
  \multicolumn{1}{c}{Southeast Asia} &
  \multicolumn{1}{c}{LATAM} &
  \textbf{\xmark} \\ \cline{2-7} 
\multicolumn{1}{|c}{} &
  \multicolumn{1}{c}{Detmold} &
  \multicolumn{1}{c}{Zagrebacka} &
  \multicolumn{1}{c}{Detmold} &
  \multicolumn{1}{c}{Southeast Asia} &
  \multicolumn{1}{c}{LATAM} &
  \textbf{\xmark} \\ \cline{2-7} 
\multicolumn{1}{|c}{} &
  \multicolumn{1}{c}{Detmold} &
  \multicolumn{1}{c}{Zagrebacka} &
  \multicolumn{1}{c}{Detmold} &
  \multicolumn{1}{c}{Southeast Asia} &
  \multicolumn{1}{c}{LATAM} &
  \textbf{\xmark} \\ \cline{2-7} 
\multicolumn{1}{|c}{} &
  \multicolumn{1}{c}{Detmold} &
  \multicolumn{1}{c}{Zagrebacka} &
  \multicolumn{1}{c}{Detmold} &
  \multicolumn{1}{c}{Southeast Asia} &
  \multicolumn{1}{c}{LATAM} &
  \textbf{\xmark} \\ \hline\hline
\multicolumn{1}{|c}{\multirow{10}{*}{\textbf{TabSyn}}} &
  \multicolumn{1}{c}{Manila} &
  \multicolumn{1}{c}{Capital Nacional} &
  \multicolumn{1}{c}{Filipinas} &
  \multicolumn{1}{c}{Southeast Asia} &
  \multicolumn{1}{c}{Pacific Asia} &
  \cellcolor{yellow}\cellcolor{yellow}\textbf{\checkmark} \\ \cline{2-7} 
\multicolumn{1}{|c}{} &
  \multicolumn{1}{c}{Tokio} &
  \multicolumn{1}{c}{Tokyo} &
  \multicolumn{1}{c}{Japan} &
  \multicolumn{1}{c}{Eastern Asia} &
  \multicolumn{1}{c}{Pacific Asia} &
  \cellcolor{yellow}\cellcolor{yellow}\textbf{\checkmark} \\ \cline{2-7} 
\multicolumn{1}{|c}{} &
  \multicolumn{1}{c}{Concord} &
  \multicolumn{1}{c}{New Hampshire} &
  \multicolumn{1}{c}{United States} &
  \multicolumn{1}{c}{East of USA} &
  \multicolumn{1}{c}{USCA} &
  \cellcolor{yellow}\cellcolor{yellow}\textbf{\checkmark} \\ \cline{2-7} 
\multicolumn{1}{|c}{} &
  \multicolumn{1}{c}{Piedecuesta} &
  \multicolumn{1}{c}{Henan} &
  \multicolumn{1}{c}{China} &
  \multicolumn{1}{c}{Eastern Asia} &
  \multicolumn{1}{c}{Pacific Asia} &
  \textbf{\xmark} \\ \cline{2-7} 
\multicolumn{1}{|c}{} &
  \multicolumn{1}{c}{Miguel Hidalgo} &
  \multicolumn{1}{c}{New Mexico} &
  \multicolumn{1}{c}{Mexico} &
  \multicolumn{1}{c}{Central America} &
  \multicolumn{1}{c}{LATAM} &
  \cellcolor{yellow}\cellcolor{yellow}\textbf{\checkmark} \\ \cline{2-7} 
\multicolumn{1}{|c}{} &
  \multicolumn{1}{c}{Lakeville} &
  \multicolumn{1}{c}{Mecklenburg-Western Pomerania} &
  \multicolumn{1}{c}{China} &
  \multicolumn{1}{c}{Eastern Asia} &
  \multicolumn{1}{c}{Pacific Asia} &
  \textbf{\xmark} \\ \cline{2-7} 
\multicolumn{1}{|c}{} &
  \multicolumn{1}{c}{Baguio City} &
  \multicolumn{1}{c}{Osjecko-Baranjska} &
  \multicolumn{1}{c}{Norway} &
  \multicolumn{1}{c}{Northern Europe} &
  \multicolumn{1}{c}{Europe} &
  \textbf{\xmark} \\ \cline{2-7} 
\multicolumn{1}{|c}{} &
  \multicolumn{1}{c}{Gaziemir} &
  \multicolumn{1}{c}{Vilnius} &
  \multicolumn{1}{c}{Iraq} &
  \multicolumn{1}{c}{Southeast Asia} &
  \multicolumn{1}{c}{Pacific Asia} &
  \textbf{\xmark} \\ \cline{2-7} 
\multicolumn{1}{|c}{} &
  \multicolumn{1}{c}{Yucaipa} &
  \multicolumn{1}{c}{Guangdong} &
  \multicolumn{1}{c}{China} &
  \multicolumn{1}{c}{Eastern Asia} &
  \multicolumn{1}{c}{Pacific Asia} &
  \textbf{\xmark} \\ \cline{2-7} 
\multicolumn{1}{|c}{} &
  \multicolumn{1}{c}{Erftstadt} &
  \multicolumn{1}{c}{Uusimaa} &
  \multicolumn{1}{c}{Italy} &
  \multicolumn{1}{c}{Southern Europe} &
  \multicolumn{1}{c}{Europe} &
  \textbf{\xmark} \\ \hline\hline
\multicolumn{1}{|c}{\multirow{10}{*}{\textbf{GReaT}}} &
  \multicolumn{1}{c}{Puebla} &
  \multicolumn{1}{c}{Puebla} &
  \multicolumn{1}{c}{Mexico} &
  \multicolumn{1}{c}{Central America} &
  \multicolumn{1}{c}{LATAM} &
  \cellcolor{yellow}\cellcolor{yellow}\textbf{\checkmark} \\ \cline{2-7} 
\multicolumn{1}{|c}{} &
  \multicolumn{1}{c}{Zhuzhou} &
  \multicolumn{1}{c}{Liaoning} &
  \multicolumn{1}{c}{China} &
  \multicolumn{1}{c}{Eastern Asia} &
  \multicolumn{1}{c}{Pacific Asia} &
  \textbf{\xmark} \\ \cline{2-7} 
\multicolumn{1}{|c}{} &
  \multicolumn{1}{c}{Nicolas Romero} &
  \multicolumn{1}{c}{Mexico} &
  \multicolumn{1}{c}{Mexico} &
  \multicolumn{1}{c}{Central America} &
  \multicolumn{1}{c}{LATAM} &
  \cellcolor{yellow}\cellcolor{yellow}\textbf{\checkmark} \\ \cline{2-7} 
\multicolumn{1}{|c}{} &
  \multicolumn{1}{c}{Munich} &
  \multicolumn{1}{c}{Bavaria} &
  \multicolumn{1}{c}{Germany} &
  \multicolumn{1}{c}{Western Europe} &
  \multicolumn{1}{c}{Europe} &
  \cellcolor{yellow}\cellcolor{yellow}\textbf{\checkmark} \\ \cline{2-7} 
\multicolumn{1}{|c}{} &
  \multicolumn{1}{c}{Seattle} &
  \multicolumn{1}{c}{Washington} &
  \multicolumn{1}{c}{United States} &
  \multicolumn{1}{c}{West of USA} &
  \multicolumn{1}{c}{USCA} &
  \cellcolor{yellow}\cellcolor{yellow}\textbf{\checkmark} \\ \cline{2-7} 
\multicolumn{1}{|c}{} &
  \multicolumn{1}{c}{Culiacan} &
  \multicolumn{1}{c}{Sinaloa} &
  \multicolumn{1}{c}{Mexico} &
  \multicolumn{1}{c}{Central America} &
  \multicolumn{1}{c}{LATAM} &
  \cellcolor{yellow}\cellcolor{yellow}\textbf{\checkmark} \\ \cline{2-7} 
\multicolumn{1}{|c}{} &
  \multicolumn{1}{c}{Vitoria} &
  \multicolumn{1}{c}{Basque Country} &
  \multicolumn{1}{c}{Spain} &
  \multicolumn{1}{c}{Southern Europe} &
  \multicolumn{1}{c}{Europe} &
  \cellcolor{yellow}\cellcolor{yellow}\textbf{\checkmark} \\ \cline{2-7} 
\multicolumn{1}{|c}{} &
  \multicolumn{1}{c}{Reims} &
  \multicolumn{1}{c}{Alsace-Champagne-Ardenne-Lorraine} &
  \multicolumn{1}{c}{France} &
  \multicolumn{1}{c}{Western Europe} &
  \multicolumn{1}{c}{Europe} &
  \cellcolor{yellow}\cellcolor{yellow}\textbf{\checkmark} \\ \cline{2-7} 
\multicolumn{1}{|c}{} &
  \multicolumn{1}{c}{Cuneo} &
  \multicolumn{1}{c}{Piedmont} &
  \multicolumn{1}{c}{Italy} &
  \multicolumn{1}{c}{Southern Europe} &
  \multicolumn{1}{c}{Europe} &
  \cellcolor{yellow}\cellcolor{yellow}\textbf{\checkmark} \\ \cline{2-7} 
\multicolumn{1}{|c}{} &
  \multicolumn{1}{c}{13551} &
  \multicolumn{1}{c}{North Rhine-Westphalia} &
  \multicolumn{1}{c}{Germany} &
  \multicolumn{1}{c}{Western Europe} &
  \multicolumn{1}{c}{Europe} &
  \textbf{\xmark} \\ \hline\hline
\multicolumn{1}{|c}{\multirow{10}{*}{\textbf{SMOTE}}} &
  \multicolumn{1}{c}{Coyoacan} &
  \multicolumn{1}{c}{Federal District} &
  \multicolumn{1}{c}{Mexico} &
  \multicolumn{1}{c}{Central America} &
  \multicolumn{1}{c}{LATAM} &
  \cellcolor{yellow}\cellcolor{yellow}\textbf{\checkmark} \\ \cline{2-7} 
\multicolumn{1}{|c}{} &
  \multicolumn{1}{c}{Lancaster} &
  \multicolumn{1}{c}{California} &
  \multicolumn{1}{c}{United States} &
  \multicolumn{1}{c}{West of USA} &
  \multicolumn{1}{c}{USCA} &
  \cellcolor{yellow}\cellcolor{yellow}\textbf{\checkmark} \\ \cline{2-7} 
\multicolumn{1}{|c}{} &
  \multicolumn{1}{c}{Wiesbaden} &
  \multicolumn{1}{c}{Hesse} &
  \multicolumn{1}{c}{Germany} &
  \multicolumn{1}{c}{Western Europe} &
  \multicolumn{1}{c}{Europe} &
  \cellcolor{yellow}\cellcolor{yellow}\textbf{\checkmark} \\ \cline{2-7} 
\multicolumn{1}{|c}{} &
  \multicolumn{1}{c}{Ciego de Avila} &
  \multicolumn{1}{c}{Ciego de Avila} &
  \multicolumn{1}{c}{Cuba} &
  \multicolumn{1}{c}{Caribbean} &
  \multicolumn{1}{c}{LATAM} &
  \cellcolor{yellow}\cellcolor{yellow}\textbf{\checkmark} \\ \cline{2-7} 
\multicolumn{1}{|c}{} &
  \multicolumn{1}{c}{Grodno} &
  \multicolumn{1}{c}{Grodno} &
  \multicolumn{1}{c}{Belarus} &
  \multicolumn{1}{c}{Eastern Europe} &
  \multicolumn{1}{c}{Europe} &
  \cellcolor{yellow}\cellcolor{yellow}\textbf{\checkmark} \\ \cline{2-7} 
\multicolumn{1}{|c}{} &
  \multicolumn{1}{c}{Bugia} &
  \multicolumn{1}{c}{Buja} &
  \multicolumn{1}{c}{Argelia} &
  \multicolumn{1}{c}{North Africa} &
  \multicolumn{1}{c}{Africa} &
  \cellcolor{yellow}\cellcolor{yellow}\textbf{\checkmark} \\ \cline{2-7} 
\multicolumn{1}{|c}{} &
  \multicolumn{1}{c}{Nagpur} &
  \multicolumn{1}{c}{Maharashtra} &
  \multicolumn{1}{c}{India} &
  \multicolumn{1}{c}{South Asia} &
  \multicolumn{1}{c}{Pacific Asia} &
  \cellcolor{yellow}\cellcolor{yellow}\textbf{\checkmark} \\ \cline{2-7} 
\multicolumn{1}{|c}{} &
  \multicolumn{1}{c}{Nacka} &
  \multicolumn{1}{c}{Stockholm} &
  \multicolumn{1}{c}{Sweden} &
  \multicolumn{1}{c}{Northern Europe} &
  \multicolumn{1}{c}{Europe} &
  \cellcolor{yellow}\cellcolor{yellow}\textbf{\checkmark} \\ \cline{2-7} 
\multicolumn{1}{|c}{} &
  \multicolumn{1}{c}{Aachen} &
  \multicolumn{1}{c}{North Rhine-Westphalia} &
  \multicolumn{1}{c}{Germany} &
  \multicolumn{1}{c}{Western Europe} &
  \multicolumn{1}{c}{Europe} &
  \cellcolor{yellow}\cellcolor{yellow}\textbf{\checkmark} \\ \cline{2-7} 
\multicolumn{1}{|c}{} &
  \multicolumn{1}{c}{Oyonnax} &
  \multicolumn{1}{c}{Auvergne-Rhone-Alpes} &
  \multicolumn{1}{c}{France} &
  \multicolumn{1}{c}{Western Europe} &
  \multicolumn{1}{c}{Europe} &
  \cellcolor{yellow}\textbf{\checkmark} \\ \hline
\end{tabular}%
}
\end{table}

\renewcommand{\arraystretch}{1.2} 
\begin{table}[t]
\centering
\caption{Temporal dependency preservation in synthetic tabular data}
\vspace{1em} 
\label{tab:temperal}
\resizebox{0.65\textwidth}{!}{%
\begin{tabular}{|cccc|}
\hline
\multicolumn{1}{|c}{\multirow{3}{*}{\textbf{Methods}}} &
  \multicolumn{2}{c}{\textbf{Group 3 (Temporal Data)}} &
  \multirow{3}{*}{\textbf{\begin{tabular}[c]{@{}c@{}}Preserved\end{tabular}}} \\ \cline{2-3}
\multicolumn{1}{|c}{}                                   & \multicolumn{1}{c}{\textbf{Attribute 1}} & \multicolumn{1}{c}{\textbf{Attribute 2}}   &              \\ \cline{2-3}
\multicolumn{1}{|c}{}                                   & \multicolumn{1}{c}{\textbf{Order Date}}  & \multicolumn{1}{c}{\textbf{Delivery Date}} &              \\ \hline\hline
\multicolumn{1}{|c}{\multirow{3}{*}{\textbf{Original Tables}}} &
  \multicolumn{1}{c}{19/08/2015  12:59:00} &
  \multicolumn{1}{c}{24/08/2015  12:59:00} &
  \cellcolor{yellow}\textbf{\checkmark} \\ \cline{2-4} 
\multicolumn{1}{|c}{}                                   & \multicolumn{1}{c}{16/06/2015  13:46:00} & \multicolumn{1}{c}{19/06/2015  13:46:00}   & \cellcolor{yellow}\textbf{\checkmark} \\ \cline{2-4} 
\multicolumn{1}{|c}{}                                   & \multicolumn{1}{c}{14/04/2016  23:41:00} & \multicolumn{1}{c}{15/04/2016  11:41:00}   & \cellcolor{yellow}\textbf{\checkmark} \\ \hline\hline
\multicolumn{1}{|c}{\multirow{10}{*}{\textbf{CTGAN}}}   & \multicolumn{1}{c}{10/07/2016  07:38:18} & \multicolumn{1}{c}{04/03/2015  04:52:46}   & \textbf{\xmark}  \\ \cline{2-4} 
\multicolumn{1}{|c}{}                                   & \multicolumn{1}{c}{14/03/2017  08:57:22} & \multicolumn{1}{c}{24/02/2015  03:31:49}   & \textbf{\xmark}  \\ \cline{2-4} 
\multicolumn{1}{|c}{}                                   & \multicolumn{1}{c}{02/08/2017  12:26:03} & \multicolumn{1}{c}{08/11/2015  04:32:00}   & \textbf{\xmark}  \\ \cline{2-4} 
\multicolumn{1}{|c}{}                                   & \multicolumn{1}{c}{14/01/2015  17:23:01} & \multicolumn{1}{c}{30/01/2015  21:37:26}   & \cellcolor{yellow}\textbf{\checkmark} \\ \cline{2-4} 
\multicolumn{1}{|c}{}                                   & \multicolumn{1}{c}{09/03/2015  03:13:50} & \multicolumn{1}{c}{21/04/2016  01:04:10}   & \cellcolor{yellow}\textbf{\checkmark} \\ \cline{2-4} 
\multicolumn{1}{|c}{}                                   & \multicolumn{1}{c}{24/08/2015  21:04:44} & \multicolumn{1}{c}{29/10/2016  00:46:55}   & \cellcolor{yellow}\textbf{\checkmark} \\ \cline{2-4} 
\multicolumn{1}{|c}{}                                   & \multicolumn{1}{c}{15/01/2017  21:17:56} & \multicolumn{1}{c}{21/05/2015  17:55:06}   & \textbf{\xmark}  \\ \cline{2-4} 
\multicolumn{1}{|c}{}                                   & \multicolumn{1}{c}{24/08/2015  12:22:23} & \multicolumn{1}{c}{13/02/2016  09:49:48}   & \cellcolor{yellow}\textbf{\checkmark} \\ \cline{2-4} 
\multicolumn{1}{|c}{}                                   & \multicolumn{1}{c}{20/03/2017  02:38:29} & \multicolumn{1}{c}{01/03/2017  19:23:58}   & \textbf{\xmark}  \\ \cline{2-4} 
\multicolumn{1}{|c}{}                                   & \multicolumn{1}{c}{26/09/2016  10:14:31} & \multicolumn{1}{c}{02/12/2015  15:52:12}   & \textbf{\xmark}  \\ \hline\hline
\multicolumn{1}{|c}{\multirow{10}{*}{\textbf{TabDDPM}}} & \multicolumn{1}{c}{31/01/2018  23:36:58} & \multicolumn{1}{c}{03/01/2015 00:00:00}    & \textbf{\xmark}  \\ \cline{2-4} 
\multicolumn{1}{|c}{}                                   & \multicolumn{1}{c}{01/01/2015 00:00:00}  & \multicolumn{1}{c}{03/01/2015 00:00:00}    & \cellcolor{yellow}\textbf{\checkmark} \\ \cline{2-4} 
\multicolumn{1}{|c}{}                                   & \multicolumn{1}{c}{01/01/2015 00:00:00}  & \multicolumn{1}{c}{03/01/2015 00:00:00}    & \cellcolor{yellow}\textbf{\checkmark} \\ \cline{2-4} 
\multicolumn{1}{|c}{}                                   & \multicolumn{1}{c}{01/01/2015 00:00:00}  & \multicolumn{1}{c}{03/01/2015 00:00:00}    & \cellcolor{yellow}\textbf{\checkmark} \\ \cline{2-4} 
\multicolumn{1}{|c}{}                                   & \multicolumn{1}{c}{01/01/2015 00:00:00}  & \multicolumn{1}{c}{06/02/2018  18:43:12}   & \cellcolor{yellow}\textbf{\checkmark} \\ \cline{2-4} 
\multicolumn{1}{|c}{}                                   & \multicolumn{1}{c}{31/01/2018  23:36:58} & \multicolumn{1}{c}{03/01/2015 00:00:00}    & \textbf{\xmark}  \\ \cline{2-4} 
\multicolumn{1}{|c}{}                                   & \multicolumn{1}{c}{01/01/2015 00:00:00}  & \multicolumn{1}{c}{06/02/2018  18:43:12}   & \cellcolor{yellow}\textbf{\checkmark} \\ \cline{2-4} 
\multicolumn{1}{|c}{}                                   & \multicolumn{1}{c}{31/01/2018  23:36:58} & \multicolumn{1}{c}{06/02/2018  18:43:12}   & \cellcolor{yellow}\textbf{\checkmark} \\ \cline{2-4} 
\multicolumn{1}{|c}{}                                   & \multicolumn{1}{c}{31/01/2018  23:36:58} & \multicolumn{1}{c}{03/01/2015 00:00:00}    & \textbf{\xmark}  \\ \cline{2-4} 
\multicolumn{1}{|c}{}                                   & \multicolumn{1}{c}{31/01/2018  23:36:58} & \multicolumn{1}{c}{06/02/2018  18:43:12}   & \cellcolor{yellow}\textbf{\checkmark} \\ \hline\hline
\multicolumn{1}{|c}{\multirow{10}{*}{\textbf{TabSyn}}} &
  \multicolumn{1}{c}{25/08/2017  02:09:36} &
  \multicolumn{1}{c}{31/08/2017  06:43:12} &
  \cellcolor{yellow}\textbf{\checkmark} \\ \cline{2-4} 
\multicolumn{1}{|c}{}                                   & \multicolumn{1}{c}{06/06/2015  04:24:58} & \multicolumn{1}{c}{13/06/2015  00:43:12}   & \cellcolor{yellow}\textbf{\checkmark} \\ \cline{2-4} 
\multicolumn{1}{|c}{}                                   & \multicolumn{1}{c}{22/10/2015  06:43:12} & \multicolumn{1}{c}{27/10/2015  05:54:14}   & \cellcolor{yellow}\textbf{\checkmark} \\ \cline{2-4} 
\multicolumn{1}{|c}{}                                   & \multicolumn{1}{c}{01/12/2016  03:00:00} & \multicolumn{1}{c}{01/12/2016  10:30:43}   & \textbf{\xmark}  \\ \cline{2-4} 
\multicolumn{1}{|c}{}                                   & \multicolumn{1}{c}{19/12/2015  08:24:00} & \multicolumn{1}{c}{16/12/2015  18:14:24}   & \textbf{\xmark}  \\ \cline{2-4} 
\multicolumn{1}{|c}{}                                   & \multicolumn{1}{c}{01/02/2017  08:42:43} & \multicolumn{1}{c}{06/02/2017  01:29:17}   & \cellcolor{yellow}\textbf{\checkmark} \\ \cline{2-4} 
\multicolumn{1}{|c}{}                                   & \multicolumn{1}{c}{17/07/2016  16:19:12} & \multicolumn{1}{c}{18/07/2016  23:16:48}   & \cellcolor{yellow}\textbf{\checkmark} \\ \cline{2-4} 
\multicolumn{1}{|c}{}                                   & \multicolumn{1}{c}{13/08/2016  11:48:29} & \multicolumn{1}{c}{10/08/2016  12:57:36}   & \textbf{\xmark}  \\ \cline{2-4} 
\multicolumn{1}{|c}{}                                   & \multicolumn{1}{c}{05/07/2015  22:33:36} & \multicolumn{1}{c}{12/07/2015  04:33:36}   & \cellcolor{yellow}\textbf{\checkmark} \\ \cline{2-4} 
\multicolumn{1}{|c}{}                                   & \multicolumn{1}{c}{07/10/2016  14:03:50} & \multicolumn{1}{c}{15/10/2016  15:28:48}   & \cellcolor{yellow}\textbf{\checkmark} \\ \hline\hline
\multicolumn{1}{|c}{\multirow{10}{*}{\textbf{GReaT}}}   & \multicolumn{1}{c}{21/07/2017  11:18:00} & \multicolumn{1}{c}{23/07/2017  11:18:00}   & \cellcolor{yellow}\textbf{\checkmark} \\ \cline{2-4} 
\multicolumn{1}{|c}{}                                   & \multicolumn{1}{c}{14/06/2016  15:22:00} & \multicolumn{1}{c}{16/06/2016  15:22:00}   & \cellcolor{yellow}\textbf{\checkmark} \\ \cline{2-4} 
\multicolumn{1}{|c}{}                                   & \multicolumn{1}{c}{23/06/2017  01:00:00} & \multicolumn{1}{c}{28/06/2017  01:00:00}   & \cellcolor{yellow}\textbf{\checkmark} \\ \cline{2-4} 
\multicolumn{1}{|c}{}                                   & \multicolumn{1}{c}{31/01/2015  10:49:00} & \multicolumn{1}{c}{02/02/2015  10:49:00}   & \cellcolor{yellow}\textbf{\checkmark} \\ \cline{2-4} 
\multicolumn{1}{|c}{}                                   & \multicolumn{1}{c}{16/10/2016  00:50:00} & \multicolumn{1}{c}{19/10/2016  00:50:00}   & \cellcolor{yellow}\textbf{\checkmark} \\ \cline{2-4} 
\multicolumn{1}{|c}{}                                   & \multicolumn{1}{c}{16/08/2017  04:36:00} & \multicolumn{1}{c}{22/08/2017  04:36:00}   & \cellcolor{yellow}\textbf{\checkmark} \\ \cline{2-4} 
\multicolumn{1}{|c}{}                                   & \multicolumn{1}{c}{03/08/2016  11:47:00} & \multicolumn{1}{c}{09/08/2016  11:47:00}   & \cellcolor{yellow}\textbf{\checkmark} \\ \cline{2-4} 
\multicolumn{1}{|c}{}                                   & \multicolumn{1}{c}{10/03/2015  22:40:00} & \multicolumn{1}{c}{13/03/2015  22:40:00}   & \cellcolor{yellow}\textbf{\checkmark} \\ \cline{2-4} 
\multicolumn{1}{|c}{}                                   & \multicolumn{1}{c}{16/01/2015  22:19:11} & \multicolumn{1}{c}{16/01/2015  10:19:00}   & \textbf{\xmark}  \\ \cline{2-4} 
\multicolumn{1}{|c}{}                                   & \multicolumn{1}{c}{21/03/2016  21:16:00} & \multicolumn{1}{c}{25/03/2016  21:16:00}   & \cellcolor{yellow}\textbf{\checkmark} \\ \hline\hline
\multicolumn{1}{|c}{\multirow{10}{*}{\textbf{SMOTE}}}   & \multicolumn{1}{c}{31/05/2016  01:07:42} & \multicolumn{1}{c}{04/06/2016  13:20:19}   & \cellcolor{yellow}\textbf{\checkmark} \\ \cline{2-4} 
\multicolumn{1}{|c}{}                                   & \multicolumn{1}{c}{21/07/2016  11:38:00} & \multicolumn{1}{c}{23/07/2016  11:38:00}   & \cellcolor{yellow}\textbf{\checkmark} \\ \cline{2-4} 
\multicolumn{1}{|c}{}                                   & \multicolumn{1}{c}{06/05/2017  01:00:54} & \multicolumn{1}{c}{08/05/2017  16:36:31}   & \cellcolor{yellow}\textbf{\checkmark} \\ \cline{2-4} 
\multicolumn{1}{|c}{}                                   & \multicolumn{1}{c}{15/10/2016  00:37:28} & \multicolumn{1}{c}{16/10/2016  22:47:19}   & \cellcolor{yellow}\textbf{\checkmark} \\ \cline{2-4} 
\multicolumn{1}{|c}{}                                   & \multicolumn{1}{c}{22/07/2017  04:52:12} & \multicolumn{1}{c}{24/07/2017  03:10:12}   & \cellcolor{yellow}\textbf{\checkmark} \\ \cline{2-4} 
\multicolumn{1}{|c}{}                                   & \multicolumn{1}{c}{10/02/2016  05:15:24} & \multicolumn{1}{c}{15/02/2016  14:55:47}   & \cellcolor{yellow}\textbf{\checkmark} \\ \cline{2-4} 
\multicolumn{1}{|c}{}                                   & \multicolumn{1}{c}{10/01/2015  00:27:21} & \multicolumn{1}{c}{14/01/2015  04:51:29}   & \cellcolor{yellow}\textbf{\checkmark} \\ \cline{2-4} 
\multicolumn{1}{|c}{}                                   & \multicolumn{1}{c}{16/01/2016  00:38:18} & \multicolumn{1}{c}{19/01/2016  09:41:35}   & \cellcolor{yellow}\textbf{\checkmark} \\ \cline{2-4} 
\multicolumn{1}{|c}{}                                   & \multicolumn{1}{c}{22/05/2015  11:33:20} & \multicolumn{1}{c}{25/05/2015  11:19:17}   & \cellcolor{yellow}\textbf{\checkmark} \\ \cline{2-4} 
\multicolumn{1}{|c}{}                                   & \multicolumn{1}{c}{17/03/2015  19:08:39} & \multicolumn{1}{c}{19/03/2015  08:27:54}   & \cellcolor{yellow}\textbf{\checkmark} \\ \hline
\end{tabular}%
}
\end{table}

\end{document}